\documentclass{article}
\usepackage[final,dblblindworkshop]{neurips_2026}
\workshoptitle{Who Verifies the Agents?}
\makeatletter
\renewcommand{\@noticestring}{%
  NeurIPS 2026 Workshop \emph{Who Verifies the Agents?}%
}
\makeatother
\usepackage[utf8]{inputenc}
\usepackage[T1]{fontenc}
\usepackage{hyperref}
\usepackage{url}
\usepackage{booktabs}
\usepackage{amsmath}
\usepackage{amssymb}
\usepackage{microtype}
\usepackage{graphicx}

\title{Who Verifies the Graph? Misspecification Attacks on\\
       Causal Action Verification for Language Agents}

\author{Fabio Rovai\\
  The Tesseract Academy\\
  London, United Kingdom\\
  \texttt{fabio@thetesseractacademy.com}}

\begin{document}
\maketitle

\begin{abstract}
Causal action verifiers gate an agent's state-changing tool calls by checking
whether each proposed intervention is identifiable against a committed
action--state graph, and they issue a certificate that carries the
identification argument and a one-sided lower confidence bound. One such
verifier, CIVeX, reports zero false executions on a confounded tool-use
benchmark. We red-team it by corrupting only the committed graph. Omitting a
single bidirected edge takes it from zero false executions to $15.3\%$ at the
benchmark's published confounding strength, with $91\%$ of its executions
harmful and utility falling from $+2.27$ to $+0.35$. Reversing one arrowhead, so
that a mediator is committed as a confounder, gives $48.9\%$ false executions
and no correct ones. Every one of these actions carries an internally valid
certificate. An attestation step that tests each observationally certified
execution against a bounded randomised sample detected both attacks, with 2
false alarms in 555 executions on a truthful graph; refusing what fails the
test, or cannot be tested, gave zero false executions in every setting we
measured. It does not restore beneficial execution: at the
published strength $97.1\%$ of beneficial actions are still never executed,
because the same misspecification rejects them before attestation runs. Those
rejections carry certificates too, and auditing them works, but its cost scales
with the number of rejections rather than the number of executions. Recovering
safety costs 127 experiments per 1,050 actions; recovering the lost value costs
614 more, at which point the audited verifier makes the honest graph's decisions
on every instance and spends exactly its experiment budget. An audit that
inspects only executions protects against wrongful action. Wrongful inaction has
to be paid for separately.
\end{abstract}

\section{Introduction}

A valid tool call is not necessarily a valid intervention. Schema validators,
policy filters and provenance checks certify the form of an action, not that
executing it will have the effect the agent expects. Causal action verifiers
close that gap by treating each proposed state-changing call as a structural
causal query over a committed action--state graph, checking identifiability, and
refusing to certify when identification fails. The certificate is an auditable
object: it names the graph, the identification argument, a point estimate, a
one-sided lower confidence bound (LCB) and a risk limit.

CIVeX \citep{rovai2026civex} is a verifier of this kind, and it is the author's
own earlier work; this paper is a red-team of it. Its reported results are
strong: zero observed false executions across a confounded tool-use benchmark,
and the only non-oracle method whose utility under a hard zero-false-execution
constraint beats an always-abstain floor. Those results hold under an
assumption the verifier does not check. The graph is committed, not learned, and
the original paper names it as the trusted computing base and lists graph
injection, adjustment-set poisoning and version drift as open concerns. That the
guarantee depends on the graph is therefore not news. What was not known is how
fast it fails, what the failures look like, and what a defence costs. This paper
measures those three things.

\begin{itemize}
\item \textbf{The published misspecification test exercises declared
ignorance.} It removes observed confounders from the graph and adds a
bidirected $T \leftrightarrow Y$ edge in their place, so the graph admits the
residual confounding. Identification then fails and the verifier routes the
affected actions to a randomised experiment or abstains. The test shows the
verifier handles a graph that admits what it does not know. It does not
exercise a graph that is wrong (Section~\ref{sec:edge}).
\item \textbf{Three misspecification attacks, measured.} Omitting the
bidirected edge, admitting a collider as a confounder, and reversing a
mediator's direction. The first and third produce certified harmful
executions; the second fails into paralysis (Sections~\ref{sec:edge},
\ref{sec:controls}).
\item \textbf{A defence, fully accounted.} Attestation against a bounded
randomised experiment detects both harmful attacks, and refusing what it flags
or cannot test gives zero observed false executions in every setting we
measured; we report the experiments spent, the disagreements found and their
cost. How the defence responds to a
detected disagreement matters as much as detecting it (Section~\ref{sec:defence}).
\item \textbf{The cost asymmetry.} Attestation of executions recovers none of
the beneficial actions the misspecification rejects. Auditing rejections does
recover them, at a cost set by the rejection volume, and full recovery costs
exactly what the honest graph spends (Section~\ref{sec:asymmetry}).
\end{itemize}

\section{Setup}
\label{sec:setup}

We run the CIVeX triage procedure, its certificate checker and the adversarial
Causal-ToolBench generator as implemented for \citet{rovai2026civex}; the only
input we change is the committed graph. Each instance is a tool-using workflow
sampled from a structural causal model with a binary treatment $T$ (the proposed
action), a continuous outcome $Y$ (workflow utility), two observed covariates
and one hidden confounder $U$ whose strength $s$ is a parameter. The hidden
confounder is adversarial: it pushes the observational estimate towards the
wrong sign. Six workflow families, seven seeds and 25 instances per family and
seed give $1{,}050$ actions per cell. Appendix~\ref{app:scm} gives the full
generating equations.

For each action the verifier searches the committed graph for a backdoor
adjustment set. If one exists, it estimates the effect by least squares on 500
observational rows with a 95\% normal-theory interval and returns EXECUTE if the
lower end of that interval is at least zero, REJECT otherwise. Both verdicts
carry a certificate. If no adjustment set exists, it returns EXPERIMENT when the
benchmark marks a safe experiment as available ($70.6\%$ of instances) and
ABSTAIN otherwise; EXPERIMENT re-runs the decision on a 500-row randomised
sample. We report the false-execution rate (share of all actions executed
although their true effect is harmful), the correct-execution rate, the share of
beneficial actions never executed, and mean utility. Utility is the benchmark's:
an execution earns $+|\tau|$ if beneficial and $-|\tau|$ if harmful, a
non-execution earns $+|\tau|$ if the action was harmful and $-0.3|\tau|$ if it
was beneficial, and every randomised experiment costs $0.05$. Because correct
refusals earn credit, the always-abstain policy scores $+0.98$ and is the floor
any useful verifier must beat.

\textbf{Threat model.} We assume an adversary, or an ordinary mistake, that can
alter the committed graph but not the verifier, the estimator, the data, or the
certificate format. The graph is the one input the verifier accepts on trust, it
is typically elicited from domain experts and stored in a registry, and all
three of our attacks are edits a careless expert could make by accident. We do
not model how such access is obtained, and we grant the defender everything
else.

Two graphs are compared throughout. The \textbf{honest} graph contains the
observed covariates as confounders and a bidirected $T \leftrightarrow Y$ edge
for the hidden confounder. The \textbf{denied} graph is identical but for that
one edge. At $s=0$ the hidden confounder has no effect, so there the denied
graph is the true one. As a reproduction check, the honest graph at the
benchmark's published strength $s=2.5$ gives $+2.27$ utility and zero false
executions, against $+2.23$ reported.

\section{Attack 1: the omitted edge}
\label{sec:edge}

\begin{figure}[t]
\centering
\includegraphics[width=\linewidth]{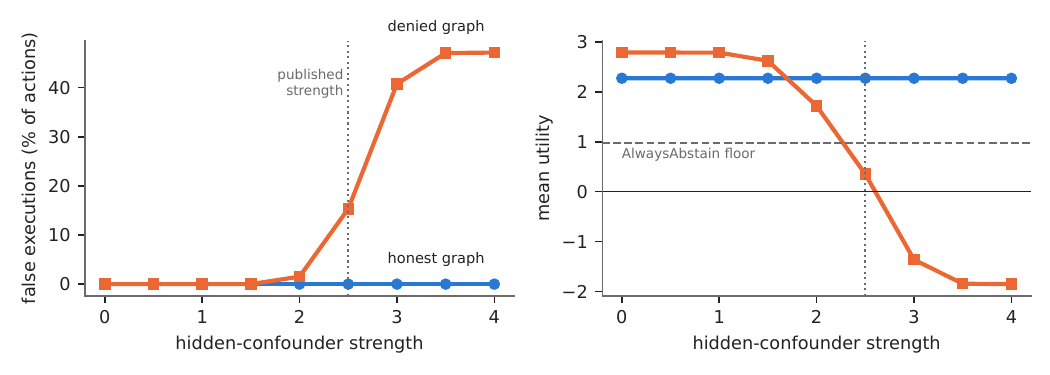}
\caption{Omitting one bidirected edge, as a function of hidden-confounder
strength. The honest graph is flat across the range. The denied graph is
harmless below strength $1.5$, falls off a cliff between $2.0$ and $3.0$, and
from $3.0$ returns negative utility. Dotted line: the benchmark's published
strength. Dashed line: the always-abstain floor. $n=1{,}050$ per point.}
\label{fig:dose}
\end{figure}

The published misspecification test relabels observed confounders as latent and
inserts a bidirected $T \leftrightarrow Y$ edge to represent the now-unblockable
path. The graph therefore stays truthful: it says that confounding exists and
cannot be adjusted away. Identification fails on exactly the affected
instances, and the verifier routes them to a randomised experiment or abstains.
Zero false executions in that test is a real result, since the experiment still
has to return the right sign, but it is a result about declared uncertainty.

The realistic failure, and the obvious injection target, is the opposite: a
graph that omits the bidirected edge and so asserts that the observed
adjustment set is sufficient. Backdoor adjustment then succeeds on an
insufficient set, the estimate absorbs the omitted confounding, the LCB is
computed on that biased estimate, and the certificate validates. Nothing in the
certificate is internally wrong.

Figure~\ref{fig:dose} sweeps the hidden confounder's strength. Up to $1.0$ the
denial is harmless and even profitable: the denied graph returns $+2.79$
against the honest graph's $+2.27$, because the honest graph pays for
experiments to guard against confounding that never changes a decision. False
executions appear at strength $2.0$ ($1.5\%$). At $2.5$, the published setting,
the rate is $15.3\%$ (per-seed range $13.3$--$17.3\%$, never zero), $91\%$ of
all executions are harmful, and utility is $+0.35$ (per seed $+0.12$ to
$+0.55$, against $+2.14$ to $+2.49$ for the honest graph). From $3.0$ upward
$41$--$47\%$ of all actions are false executions, every execution is harmful,
and utility is negative, $-1.36$ at $3.0$ and $-1.85$ at $4.0$, far below the
always-abstain floor. The honest line does not move anywhere in the sweep.
Whether the omitted edge is harmless or catastrophic depends on a quantity the
graph has declared absent, and the certificate looks the same on both sides of
the cliff.

\section{Attacks 2 and 3: bad controls and a reversed arrowhead}
\label{sec:controls}

\begin{figure}[t]
\centering
\includegraphics[width=\linewidth]{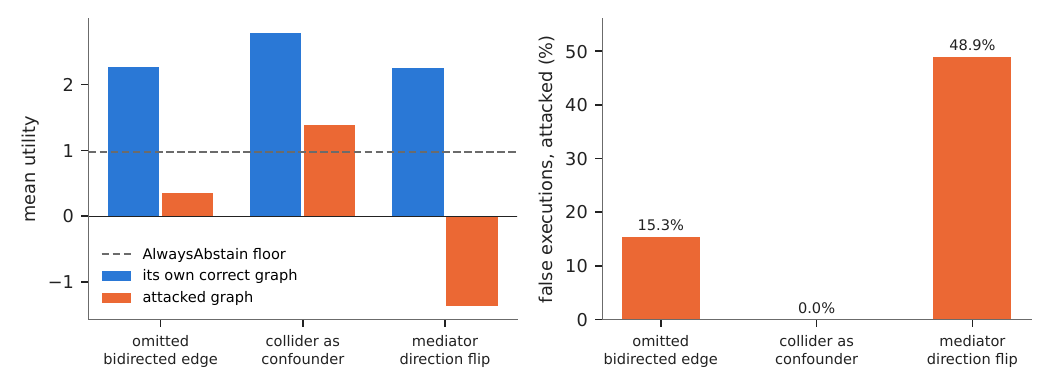}
\caption{Three misspecifications, each against its own correct-graph control
(the three run on different data-generating regimes, so a single shared baseline
would mislead). Bad controls fail into paralysis; the omitted edge and the
reversed arrowhead fail into wrongful action. $n=1{,}050$ per bar.}
\label{fig:taxonomy}
\end{figure}

\textbf{A collider admitted as a confounder.} On a regime with no hidden
confounding, where the graph is otherwise exactly true, we synthesise
$C = 0.8\,T + 0.8\,(Y - \bar Y) + \varepsilon$ and commit it as a confounder.
Conditioning on a descendant of the outcome cuts correct executions from
$52.9\%$ to $7.9\%$ (per seed $3.3$--$11.3\%$) and utility from $+2.79$ to
$+1.39$, with zero false executions. In this regime the bad control fails
safe: the verifier stops acting rather than acting wrongly.

\textbf{A mediator committed backwards.} We build an SCM with a genuine
mediator, $T \rightarrow M \rightarrow Y$ alongside a direct $T \rightarrow Y$
path, in which the direct effect carries the opposite sign to the total effect.
The correct graph excludes $M$ from the adjustment set, because the backdoor
search rejects descendants of the treatment, and recovers the total effect. The
attacked graph asserts $M \rightarrow T$ instead of $T \rightarrow M$: a
direction error on a variable that genuinely exists, with nothing omitted and
nothing fabricated. $M$ then enters the adjustment set, the indirect path is
blocked, and the estimate recovers the direct effect, whose sign is inverted.

\begin{table}[t]
\centering
\small
\begin{tabular}{lrrrr}
\toprule
committed graph & false exec & correct exec & utility & beneficial lost \\
\midrule
correct ($T \rightarrow M$) & $0.0\%$ & $51.1\%$ & $+2.26$ & $0.0\%$ \\
direction flip ($M \rightarrow T$) & $48.9\%$ & $0.0\%$ & $-1.36$ & $100\%$ \\
\bottomrule
\end{tabular}
\caption{One reversed arrowhead inverts the verifier. Per-seed false-execution
range under the flip: $42.7$--$57.3\%$. $n=1{,}050$ per row.}
\label{tab:overcontrol}
\end{table}

Table~\ref{tab:overcontrol} gives the result: total inversion. Every execution
is harmful, every beneficial action is rejected, and utility is negative. The
variable is real, the graph is complete, no assumption is left undeclared, and
the identification argument is internally valid for the graph as committed. The
graph is wrong by one edge direction.

\section{A defence: attestation against a bounded experiment}
\label{sec:defence}

``No unmeasured confounding'' cannot be tested from observational data alone,
so a defence has to buy the test with an intervention. We add an attestation
step. Whenever the verifier reaches EXECUTE by observational identification,
the step draws the 500-row randomised sample, computes the difference in means
with a 95\% normal-theory interval, and compares it with the interval in the
certificate. If the two intervals are disjoint, the committed graph is declared
falsified for that action. We compare two responses to falsification:
\textbf{refuse} (abstain) and \textbf{fallback} (re-decide on the randomised
sample). Each attestation costs one experiment, $0.05$ utility, and every
utility figure below includes that cost. The test is a discrepancy diagnostic,
not a calibrated hypothesis test: overlapping intervals do not show that the
graph is correct, and we make no claim about its false-acceptance rate beyond
what we measure.

\begin{table}[t]
\centering
\small
\begin{tabular}{rrrrrrrr}
\toprule
& \multicolumn{2}{c}{attestation} & \multicolumn{2}{c}{false exec} & \multicolumn{3}{c}{utility} \\
\cmidrule(lr){2-3}\cmidrule(lr){4-5}\cmidrule(lr){6-8}
strength & experiments & falsified & none & attest & none & refuse & fallback \\
\midrule
$0.0$ & 555 & 2   & $0.0\%$  & $0.0\%$ & $+2.79$ & $+2.76$ & $+2.76$ \\
$0.5$ & 555 & 79  & $0.0\%$  & $0.0\%$ & $+2.79$ & $+2.52$ & $+2.76$ \\
$1.0$ & 554 & 547 & $0.0\%$  & $0.0\%$ & $+2.79$ & $+1.05$ & $+2.76$ \\
$1.5$ & 474 & 474 & $0.0\%$  & $0.0\%$ & $+2.62$ & $+1.03$ & $+2.60$ \\
$2.0$ & 201 & 201 & $1.5\%$  & $0.0\%$ & $+1.72$ & $+1.05$ & $+1.77$ \\
$2.5$ & 177 & 177 & $15.3\%$ & $0.0\%$ & $+0.35$ & $+1.05$ & $+1.12$ \\
$3.0$ & 428 & 428 & $40.8\%$ & $0.0\%$ & $-1.36$ & $+1.04$ & $+1.04$ \\
$4.0$ & 495 & 495 & $47.1\%$ & $0.0\%$ & $-1.85$ & $+1.03$ & $+1.03$ \\
\bottomrule
\end{tabular}
\caption{Attestation on the denied graph, itemised. Experiments: observational
EXECUTEs that were attested (one experiment each, out of $1{,}050$ actions).
Falsified: attestations whose intervals were disjoint. False executions are
shares of all actions. Utility includes the experiment cost. On the honest graph
attestation is never invoked: it has zero observational EXECUTEs at every
strength, because its bidirected edge sends every action to the experiment path,
where it already spends 741 experiments. This table assumes, as the benchmark's
experiment path does not, that a randomised sample exists for every attested
action; see the text for the case where it does not.}
\label{tab:defence}
\end{table}

Table~\ref{tab:defence} itemises the defence on the denied graph. Five things
follow from it.

\emph{It detects every harmful execution we generated.} From strength $2.0$
upward every attested execution was falsified, and false executions went to
zero at every strength. Zero observed events in $1{,}050$ actions bounds the
true rate below about $0.29\%$ at 95\% confidence by the rule of three
\citep{hanley1983if}; that is the strength of the claim, not a guarantee.

\emph{It can fail, and on a true graph it rarely does.} At strength $0$ the
denied graph is correct, and attestation flagged 2 of 555 executions ($0.4\%$).
Its cost there is almost entirely the experiments themselves: utility falls
from $+2.79$ to $+2.76$.

\emph{It is ``free on honest graphs'' only in a vacuous sense.} Our submitted
version reported a block rate of zero on the honest graph. The accounting
shows why: the honest graph never produces an observational execution, so
attestation never runs. That graph is not cheap; it spends 741 experiments per
$1{,}050$ actions.

\emph{The response matters more than the detection.} With 500 randomised rows
the test is sensitive enough to flag bias that changes no decision: at
strength $1.0$ it falsified 547 of 554 executions although the denial there
costs nothing. Refusing on falsification then destroys the verifier (utility
$+1.05$, $98.7\%$ of beneficial actions lost), while fallback keeps it
($+2.76$). A defence deployed with the refuse response would be worse than no
defence across the whole harmless range.

\emph{Availability decides whether zero is reachable.} The benchmark marks a
safe experiment available for only $70.6\%$ of actions, and
Table~\ref{tab:defence} ignores that flag. If the defender honours it and
refuses executions it cannot attest, false executions stay at zero, but on the
true graph utility drops from $+2.79$ to $+2.29$, because the 158 of 555
correct executions with no experiment available become refusals. If instead unattestable executions go
through, false executions return: $4.3\%$ at strength $2.5$ (per seed
$2.0$--$6.7\%$) and $12.4\%$ at $3.0$. Zero false executions needs either an
experiment for every eligible action or a willingness to refuse the ones
without.

\textbf{The direction flip.} Attestation also detects the reversed arrowhead:
with an experiment for every action, all 513 attacked executions were
falsified, against 2 of 537 on the correct mediator graph. Refusing then gives
zero false executions (utility $+0.58$ with availability honoured,
Table~\ref{tab:mediator}).
Fallback does not, and the reason is instructive. Our fallback re-decides on
the randomised sample \emph{using the committed graph}, which still puts $M$ in
the adjustment set. Randomising $T$ removes confounding, but it does not undo an
adjustment that blocks the causal path, so the fallback reproduces the error:
$48.9\%$ false executions with an experiment for every action, $34.6\%$ when
unattestable executions are refused. A fallback that ignores the committed
adjustment set and uses the raw difference in means brings false executions to
zero (Table~\ref{tab:mediator}). The general lesson is that a defence built for
one kind of graph error can silently inherit another.

\begin{table}[t]
\centering
\small
\begin{tabular}{llrrrr}
\toprule
graph & defence & false exec & correct exec & utility & beneficial lost \\
\midrule
correct & none & $0.0\%$ & $51.1\%$ & $+2.26$ & $0.0\%$ \\
correct & attest / fallback & $0.0\%$ & $34.4\%$ & $+1.69$ & $32.8\%$ \\
flipped & none & $48.9\%$ & $0.0\%$ & $-1.36$ & $100\%$ \\
flipped & attest / refuse & $0.0\%$ & $0.0\%$ & $+0.58$ & $100\%$ \\
flipped & attest / fallback, committed graph & $34.6\%$ & $0.0\%$ & $-0.81$ & $100\%$ \\
flipped & attest / fallback, graph-free & $0.0\%$ & $0.0\%$ & $+0.58$ & $100\%$ \\
flipped & graph-free + audit all rejections & $0.0\%$ & $34.4\%$ & $+1.68$ & $32.8\%$ \\
\bottomrule
\end{tabular}
\caption{Attestation against the direction flip, honouring experiment
availability (unattestable executions refused). Detection works; the fallback
must not reuse the committed adjustment set. The last row is the rejection audit
of Section~\ref{sec:asymmetry}. $n=1{,}050$ per row.}
\label{tab:mediator}
\end{table}

\section{The cost asymmetry}
\label{sec:asymmetry}

\begin{figure}[t]
\centering
\includegraphics[width=\linewidth]{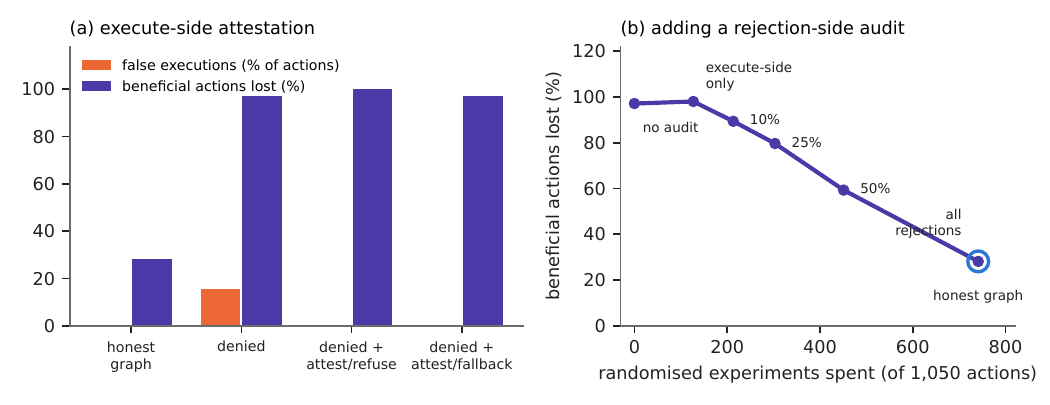}
\caption{Published confounding strength. (a) Attestation of executions, with an
experiment assumed for every attested action, removes the false executions and
leaves the lost beneficial actions where they were ($97.1\%$ with fallback,
$100\%$ with refuse). (b) Auditing a share of the
rejections as well, with experiment availability honoured: beneficial actions
come back roughly in proportion to experiments spent, and auditing every
rejection lands on the honest graph (ringed). $n=1{,}050$ per point.}
\label{fig:asym}
\end{figure}

\begin{table}[t]
\centering
\small
{\setlength{\tabcolsep}{4.5pt}\begin{tabular}{llrrrrr}
\toprule
graph & defence & experiments & false exec & correct exec & utility & beneficial lost \\
\midrule
honest & none & 741 & $0.0\%$ & $38.0\%$ & $+2.27$ & $28.1\%$ \\
denied & none & 0 & $15.3\%$ & $1.5\%$ & $+0.35$ & $97.1\%$ \\
denied & attest / refuse$^\dagger$ & 177 & $0.0\%$ & $0.0\%$ & $+1.05$ & $100.0\%$ \\
denied & attest / fallback$^\dagger$ & 177 & $0.0\%$ & $1.5\%$ & $+1.12$ & $97.1\%$ \\
denied & attest / fallback & 127 & $0.0\%$ & $1.0\%$ & $+1.10$ & $98.0\%$ \\
denied & \quad + audit 10\% of rejections & 213 & $0.0\%$ & $5.6\%$ & $+1.24$ & $89.4\%$ \\
denied & \quad + audit 25\% & 303 & $0.0\%$ & $10.8\%$ & $+1.40$ & $79.6\%$ \\
denied & \quad + audit 50\% & 451 & $0.0\%$ & $21.5\%$ & $+1.74$ & $59.3\%$ \\
denied & \quad + audit all & 741 & $0.0\%$ & $38.0\%$ & $+2.27$ & $28.1\%$ \\
\bottomrule
\end{tabular}}
\caption{Safety is cheap to recover and value is not. Published confounding
strength. Experiments are randomised samples drawn per $1{,}050$ actions.
$^\dagger$Assumes a randomised sample exists for every attested action; the
other denied rows honour the benchmark's availability flag and refuse
executions they cannot attest.}
\label{tab:asym}
\end{table}

Table~\ref{tab:asym} and Figure~\ref{fig:asym} give the paper's main result.
Attestation of executions takes false executions to zero and lifts utility from
$+0.35$ to $+1.12$, but the gain comes entirely from avoiding harm. Beneficial
execution is not recovered: $97.1\%$ of beneficial actions are lost with
fallback, exactly as without the defence, and $100\%$ with refuse, against
$28.1\%$ under the honest graph. At strength $3.0$ and above the denied graph
loses every beneficial action with or without attestation.

The mechanism is visible in the verdicts. Of the 555 beneficial actions at the
published strength, 539 are rejected at the first triage by observational
identification, with a certificate whose biased estimate has a negative lower
bound. Attestation of executions never inspects them. The fault is in the audit
path, not in the certificate: CIVeX attaches the graph, adjustment set, estimate
and interval to a REJECT exactly as it does to an EXECUTE, so a rejection can be
tested the same way. Our submitted version said that a rejected action emits no
certificate. For this verifier that was wrong.

Auditing rejections works. Applying the same attestation and fallback to a
random share of the rejections for which an experiment is available recovers
beneficial actions roughly in proportion to the share audited, with no false
executions at any budget (Table~\ref{tab:asym}, Figure~\ref{fig:asym}b). The
price is set by volume. Under the denied graph, 873 of the $1{,}050$ actions are
observational rejections, and 334 of those are correct. Recovering safety took
127 experiments; recovering the value took 614 more, $4.8$ times as many. At
strength $3.0$ the split is 298 and 443. Auditing every rejection that can be
audited reproduces the honest graph exactly: on every instance at every
strength from $1.5$ to $4.0$ it makes the honest graph's execute decisions, and
it spends the same 741 experiments. This is an identity rather than a
coincidence. At those strengths every audited verdict is falsified, so every
action with an experiment available is re-decided on the randomised sample,
which is what the honest graph's bidirected edge does from the start.

So the value was always bought with experiments, and omitting the edge hides
the need for them. The asymmetry is a property of where the audit budget goes,
not a property of certificates. An audit that inspects only executions sees the
actions that would change the world, which are the ones whose harm can later
surface in outcomes. A wrongly rejected action changes no state and leaves no
outcome to observe, so nothing downstream flags the loss, and finding it means
paying to re-examine decisions most of which were correct. A deployment that
monitors false executions alone will see a verifier with a clean record while
it quietly does nothing.

\section{Related work}

\textbf{Identification.} The certificate machinery rests on do-calculus and the
Shpitser--Pearl ID algorithm \citep{pearl2009causality, shpitser2006identification},
with backdoor and frontdoor adjustment \citep{rosenbaum1983central} as the usual
arguments. Our attacks do not challenge that theory; they hold it fixed and
corrupt its input.

\textbf{High-confidence deployment.} The decision rule ``compute a one-sided
lower bound on the effect and act only if it clears a threshold, otherwise
return no solution'' predates its use in agent verification: it is
high-confidence off-policy evaluation and improvement
\citep{thomas2015hcope, thomas2015hcpi} and the Seldonian framework
\citep{thomas2019preventing}, whose signature behaviour is returning
\textsc{no solution found} when the safety constraint cannot be certified. The
CIVeX paper cites constrained MDPs, shielded RL and safe-exploration surveys
\citep{altman1999constrained, achiam2017constrained, alshiekh2018safe,
garcia2015comprehensive} but not this branch, which is its nearest ancestor; we
correct that omission here. The difference is granularity and auditability: a
per-action gate on an off-the-shelf agent with an inspectable certificate,
rather than a bound on a learned policy. Our results say that this is also where
the exposure lies, because a per-action gate that is audited only on its
executions leaves its rejections unexamined.

\textbf{Sensitivity analysis.} Bounding the effect of unmeasured confounding
\citep{rosenbaum1983central, vanderweele2017sensitivity, cinelli2020making,
manski2003partial} is the classical response to exactly the assumption we
attack. Separately, \citet{cinelli2024controls} catalogue the good and bad
control choices that our second and third attacks instantiate: a collider
admitted as a confounder and a mediator committed backwards are both entries in
that taxonomy, and our contribution is to price them inside a verifier. A
verifier that reported a sensitivity interval alongside the LCB would degrade
more gracefully under the omitted edge. It would not help with the direction
flip, which is a structural error rather than a confounding one, and it would
not by itself recover wrongly rejected actions.

\textbf{Verifier robustness.} Work on reward hacking and specification gaming
\citep{amodei2016concrete, skalse2022reward} asks whether an agent exploits its
evaluator. We ask the adjacent question of whether the evaluator's own trusted
inputs can be corrupted, an attack surface that grows as verification moves
from scalar rewards to structured, assumption-carrying artefacts.

\section{Limitations}

The benchmark is synthetic by construction, which is what makes ground-truth
causal effects available at all; it demonstrates that observational and
interventional signs can disagree, and does not claim to match any specific
production system. The collider and mediator regimes are built to expose their
failure modes, and we sweep strength only for the omitted edge; how often such
errors occur in elicited graphs is outside what we measure. Our attacks edit the
graph directly rather than through an attacker with capabilities and a budget;
we measure the damage a corrupted graph does, not the difficulty of corrupting
one. Attestation assumes that a bounded randomised sample exists and is cheap,
which fails for irreversible actions, where verification matters most, and the
disjoint-interval test carries no calibrated error guarantee. The rejection
audit samples rejections uniformly; a targeted audit, for example of rejections
whose certified bound lies near zero, might recover value more cheaply, and we
have not tested one. We do not evaluate on a real tool-agent benchmark, so the
numbers characterise one verifier and its benchmark rather than deployed agents,
and whether other verification schemes show the same split between the cost of
safety and the cost of value is untested.

\section{Conclusion}

A causal action verifier certifies the actions it is asked to permit against a
graph it cannot check. One omitted edge or one reversed arrowhead makes every
certified execution wrong, and the certificates stay internally valid. A
bounded-experiment attestation, refusing what it flagged or could not test,
removed every false execution we generated, at a cost proportional to the
number of executions. It recovered none of the
beneficial actions the same error rejects. Those can be recovered too, by
auditing rejections, but the bill scales with the number of rejections, and
recovering all of them costs what the honest graph would have spent in the first
place. Verification work on agents should treat the committed graph as an attack
surface with its own integrity requirements, report wrongful inaction alongside
wrongful action, and budget for auditing both.

\section*{Acknowledgements}

The four reviewers and the area chair of the \emph{Who Verifies the Agents?}
workshop asked for the attestation accounting and challenged the scope of the
asymmetry claim; Sections~\ref{sec:defence} and~\ref{sec:asymmetry} are the
result.

\bibliographystyle{plainnat}
\bibliography{references}

\appendix

\section{Benchmark, verifier and defence specification}
\label{app:scm}

\textbf{Omitted-edge and collider regimes.} Each instance draws a label
$b \sim \mathrm{Bernoulli}(0.5)$ and an effect
$\tau \sim \mathcal{N}(\mu_b, 0.4^2)$, where $(\mu_{\text{beneficial}},
\mu_{\text{harmful}})$ is family-specific: $(+3.0,-3.0)$, $(+2.5,-3.0)$,
$(+3.0,-3.5)$, $(+2.0,-2.5)$, $(+2.5,-3.0)$ and $(+2.0,-3.5)$ for the six
families. The ground-truth label is $\tau > 0$. Observational data are $n=500$
rows of
\begin{align*}
X_1 &\sim \mathcal{N}(10, 2^2), \quad X_2 \sim \mathcal{N}(5, 1^2), \quad
U \sim \mathcal{N}(0,1), \\
T &\sim \mathrm{Bernoulli}\!\left(\sigma\!\left(0.3\,\tfrac{X_1-10}{2} - 0.2\,\tfrac{X_2-5}{1} + \alpha\,U\right)\right), \\
Y &= \tau T + 0.1\,X_1 + 0.05\,X_2 + \beta\,U + \varepsilon, \quad \varepsilon \sim \mathcal{N}(0,1),
\end{align*}
with $(\alpha,\beta) = (-s,+s)$ for beneficial and $(+s,+s)$ for harmful
instances, so the hidden confounder biases the observational contrast towards
the wrong sign; $s = 2.5$ is the published setting. The randomised sample keeps
the same covariates, $U$ and noise, and draws $T \sim \mathrm{Bernoulli}(0.5)$
independently of everything. A safe experiment is available for an instance
with probability $0.7$ ($70.6\%$ realised). The collider is
$C = 0.8\,T + 0.8\,(Y - \bar Y) + \mathcal{N}(0, 0.5^2)$, committed with edges
$C \rightarrow T$ and $C \rightarrow Y$, on the $s=0$ regime.

\textbf{Mediator regime.} Same covariates, no hidden confounder,
$M = m_t T + \mathcal{N}(0, 0.5^2)$ and $Y = d\,T + m_y M + 0.1\,X_1 + 0.05\,X_2
+ \varepsilon$, with $(d, m_t, m_y) = (-0.8, 1.0, 3.3)$ for beneficial actions
(total effect $+2.5$) and $(1.0, 1.0, -3.0)$ for harmful ones (total $-2.0$).
The direct effect always has the opposite sign to the total.

\textbf{Verifier.} Backdoor sets are searched on the committed graph. The
estimate is the coefficient on $T$ in an ordinary least-squares regression of
$Y$ on $T$ and the adjustment set, with standard error from the residual
variance and interval estimate $\pm 1.96$ standard errors. EXECUTE requires the
lower end to be at least $0$ and the action's risk bound ($0.05$) to be at most
$0.5$; an identified action that fails this is REJECTed with its certificate.
An unidentified action goes to EXPERIMENT if a safe experiment is available and
to ABSTAIN otherwise; EXPERIMENT re-runs the decision on the randomised sample
with the observed-covariate graph.

\textbf{Attestation and audit.} The interventional interval is the difference
in means on the randomised sample, $\pm 1.96$ Welch standard errors. The graph
is falsified for an action when this interval and the certificate's interval
are disjoint. Fallback with the committed graph re-runs the decision on the
randomised sample with the committed graph; graph-free fallback executes if the
interventional lower end is at least $0$ and rejects if its upper end is below
$0$. The rejection audit draws one uniform number per instance, fixed per seed,
and audits an observational REJECT with an available experiment when that
number falls below the budget, so the audited sets are nested across budgets.
Every experiment, whether triage, attestation or audit, costs $0.05$ utility.

\textbf{Code state.} All results come from one working copy of the CIVeX
repository. The camera-ready tables were produced by a single script that
triages each instance once, stores its decision, certificate interval,
availability and attestation outcome, and derives every policy from that
table, so all policies see identical instances. It reproduces the submitted
version's attestation results exactly (for example $+1.121$ and $97.1\%$ for
fallback at $s=2.5$).

\section{Per-seed variability}
\label{app:seeds}

\begin{table}[h]
\centering
\small
{\setlength{\tabcolsep}{4.5pt}\begin{tabular}{lrrrr}
\toprule
cell ($n=1{,}050$, 7 seeds) & false exec & per-seed range & utility & per-seed range \\
\midrule
honest graph, $s=2.5$ & $0.0\%$ & $0.0$--$0.0\%$ & $+2.27$ & $+2.14$ to $+2.49$ \\
denied graph, $s=2.0$ & $1.5\%$ & $0.7$--$2.0\%$ & $+1.72$ & $+1.57$ to $+1.92$ \\
denied graph, $s=2.5$ & $15.3\%$ & $13.3$--$17.3\%$ & $+0.35$ & $+0.12$ to $+0.55$ \\
denied graph, $s=3.0$ & $40.8\%$ & $37.3$--$44.7\%$ & $-1.36$ & $-1.51$ to $-1.17$ \\
denied + attest/fallback, $s=2.5$ & $0.0\%$ & $0.0$--$0.0\%$ & $+1.12$ & $+0.88$ to $+1.26$ \\
denied + attest/fallback, pass-through, $s=2.5$ & $4.3\%$ & $2.0$--$6.7\%$ & $+0.92$ & $+0.69$ to $+1.15$ \\
mediator direction flip & $48.9\%$ & $42.7$--$57.3\%$ & $-1.36$ & $-1.47$ to $-1.28$ \\
\bottomrule
\end{tabular}}
\caption{Seed-level spread for the headline cells.}
\label{tab:seeds}
\end{table}

\end{document}